%% file: acl2023.tex
\pdfoutput=1
\documentclass[11pt]{article}

\input{_package}

\input{_definition}

\title{Autonomous Workflow for Multimodal Fine-Grained Training Assistants Towards Mixed Reality}

\input{_author}

\begin{document}
\maketitle

\vspace*{0.5\baselineskip}
\input{0_abstract}
\input{1_introduction}

\input{2_related-work}
\input{3_workflow}

\input{4_dataset}

\input{5_experimental_setup}
\input{6_outcomes}
\input{7_discussion}
\input{8_conclusion}

% \newpage
\section*{Reproducibility}
We release resources including the source code and dataset at \url{https://github.com/Jiahuan-Pei/AutonomousDialogAgent4AugmentedReality}.

\section*{Limitations}
The generation of user requirements and the dataset relies solely on the simulation process. 
This workflow serves as a fast solution to verify the concept of an \ac{LLM} agent aiding in a specific use case, such as a LEGO assembly assistant.
However, we acknowledge that the study of user requirements are valuable and needed to build up user-centric \ac{AI} agents and \ac{MR} applications.
Besides, the demonstration codes do not optimize \ac{LLM} and \ac{VLM} simultaneously, potentially leading to suboptimal outcomes.
We have only assessed \acp{LLM} as benchmarks. However, we have not conducted separate assessments of the influence on the vision-language agent and user experience in \ac{MR}. 
We plan to explore these aspects in future work.
\section*{Ethics Statement}
We realize that there are risks in developing a large language model for users, so it is necessary to pay attention to the ethical issues.
Therefore, we use the open-resourced \acp{LLM} as benchmarks and consider user-centric points:
A user will first be provided with an explanation of what will be happening during their \ac{MR} training experience.
Users will then be provided with relevant consent forms to sign, and after signing they will be fitted with \ac{AR} glasses and the training scenario will begin.
After launching the application, the user will be greeted by the virtual assistant and prompted to confirm they would like to begin training. 
After confirming, the user will then be asked by the virtual assistant which difficulty level they would like to be trained on. 

\input{_ack}

\bibliography{acl2023}
\bibliographystyle{acl_natbib}

\appendix
\input{_appendix}

% \input{vox}

\end{document}

%% file: _package.tex
\usepackage{authblk}
\usepackage[final]{acl2023}

\usepackage{times}
\usepackage{latexsym}

\usepackage[T1]{fontenc}
\usepackage[utf8]{inputenc}

\usepackage{microtype}

\usepackage{inconsolata}
\usepackage{booktabs}
\usepackage{multirow}
\usepackage{enumerate}
\usepackage[inline]{enumitem}
\usepackage{amsfonts} 
\usepackage{amsmath,amsthm}
\usepackage{amssymb}
\usepackage{bbding}
\usepackage{graphicx}
\usepackage{subcaption}
\usepackage{hyperref} % for footnote
\usepackage{enumitem} % for inline enumitem

\setlist[itemize]{leftmargin=*, nosep}

%% file: _definition.tex
\usepackage{acronym}
\acrodef{NLP}{natural language processing}
\acrodef{HCI}{human computer interaction}

\acrodef{XR}{extended reality}
\acrodef{MR}{mixed reality}
\acrodef{AR}{augmented reality}
\acrodef{VR}{virtual reality}
\acrodef{MRTA}{extended reality training assistant}
\acrodef{QA}{question answering}
\acrodef{VQA}{vision question answering}

\acrodef{NLU}{natural language understanding}
\acrodef{DPL}{dialogue policy learning}
\acrodef{NLG}{natural language generation}
\acrodef{LLM}{large language model}
\acrodef{MIS}{multimodal immersive system}
\acrodef{DA}{dialogue agent}
\acrodef{AI}{artificial intelligence}

\acrodef{LLM}{large language model}
\acrodef{VLM}{vision-language model}
\acrodef{PEFT}{parameter-efficient fine-tuning}
\acrodef{LoRA}{low rank adaptation}
\acrodef{API}{application programming interface}

\newcommand{\OurData}{LEGO-\acs{MRTA}}

%% file: _author.tex
\author[1]{Jiahuan Pei}
\author[1]{Irene Viola}
\author[1]{Haochen Huang}
\author[3]{Junxiao Wang}
\author[1]{Moonisa Ahsan}
\author[4]{Fanghua Ye}
\author[5]{\\Yiming Jiang}
\author[5]{Yao Sai}
\author[3]{Di Wang}
\author[5]{Zhumin Chen}
\author[,5]{Pengjie Ren\thanks{~~Corresponding author.}\hspace*{0.3em}}
\author[1,2]{Pablo Cesar}
\affil[1]{\small{
Centrum Wiskunde \& Informatica, Amsterdam, The Netherlands,
\texttt{\{j.pei,irene,p.s.cesar\}@cwi.nl}
}}
\affil[2]{\small{TU Delft, Delft, The Netherlands}}
\affil[3]{\small{King Abdullah University of Science and Technology, Thuwal, Saudi Arabia}}
\affil[4]{\small{University College London, London, United Kingdom, \texttt{fanghua.ye.19@ucl.ac.uk}}}
\affil[5]{\small{
Shandong University, Qingdao, China, 
\texttt{jay.ren@outlook.com}
}}

%% file: 0_abstract.tex
\begin{abstract}
Autonomous \ac{AI} agents have emerged as promising protocols for automatically understanding the language-based environment, particularly with the exponential development of \acp{LLM}.
However, a fine-grained, comprehensive understanding of multimodal environments remains under-explored. 
%%%
This work designs an autonomous workflow tailored for integrating \ac{AI} agents seamlessly into \ac{MR} applications for fine-grained training.
We present a demonstration of a multimodal fine-grained training assistant for LEGO brick assembly in a pilot \ac{MR} environment.
Specifically, we design a cerebral language agent that integrates \acp{LLM} with memory, planning, and interaction with \ac{MR} tools and a vision-language agent, enabling agents to decide their actions based on past experiences.
Furthermore, we introduce \OurData{}, a multimodal fine-grained assembly dialogue dataset synthesized automatically in the workflow served by a commercial \ac{LLM}.
This dataset comprises multimodal instruction manuals, conversations, \ac{MR} responses, and \acl{VQA}.
Last, we present several prevailing open-resource \acp{LLM} as benchmarks, assessing their performance with and without fine-tuning on the proposed dataset.
%%%
We anticipate that the broader impact of this workflow will advance the development of smarter assistants for seamless user interaction in \ac{MR} environments, fostering research in both \ac{AI} and \acs{HCI} communities.
\end{abstract}

%% file: 1_introduction.tex
\section{Introduction}
\label{sec:introduction}

\begin{figure}[t!]
\vspace*{1.5\baselineskip}
\centering
\begin{minipage}{\columnwidth}
    \begin{subfigure}[b]{\columnwidth}
     \centering
     \includegraphics[width=\columnwidth, trim={0 5cm 0 0},clip]{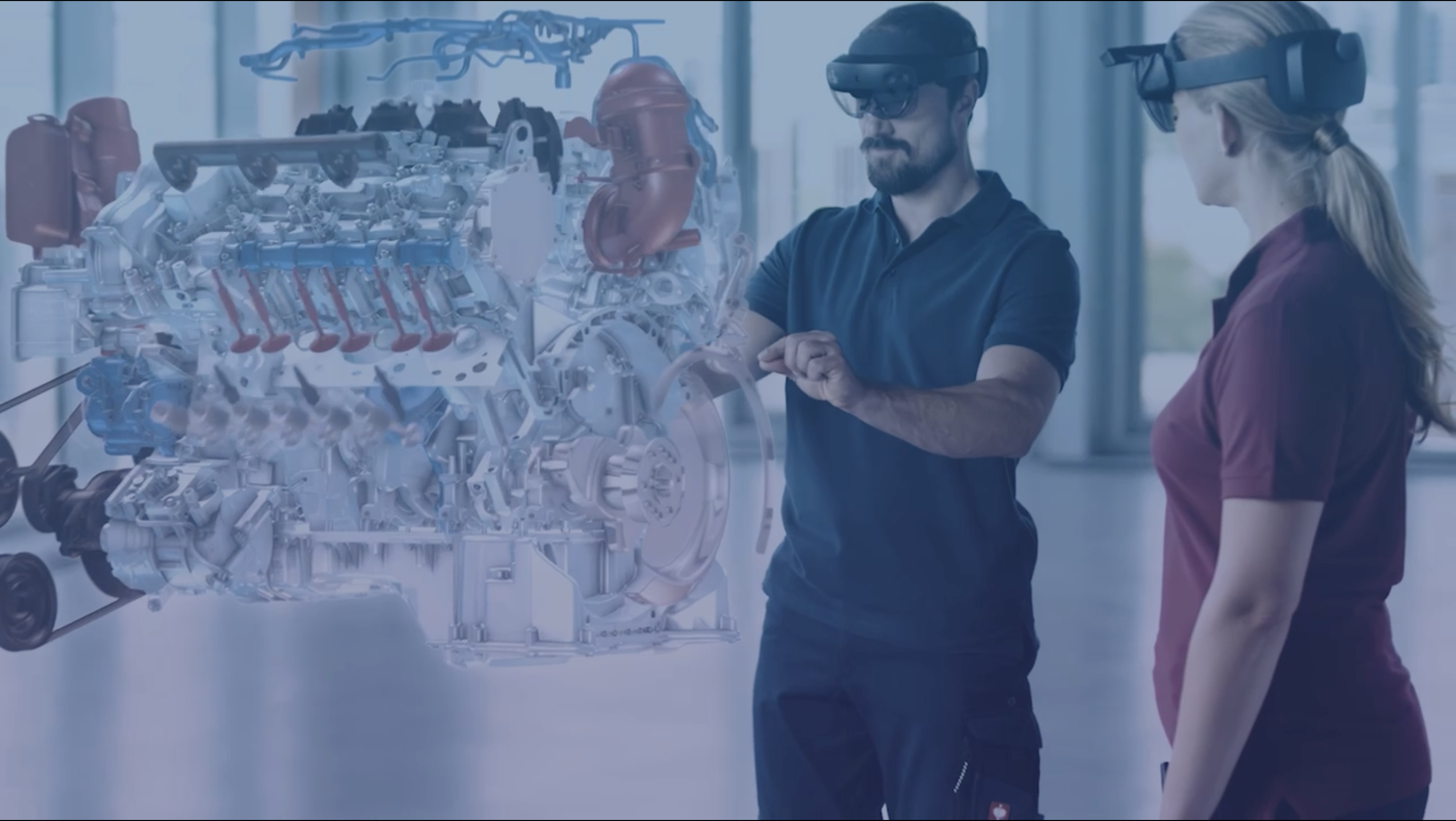}
     \caption{Industrial Car Assembly.}
    \end{subfigure}
    \hfill
    \begin{subfigure}[b]{\columnwidth}
     \centering
     \includegraphics[width=\columnwidth]{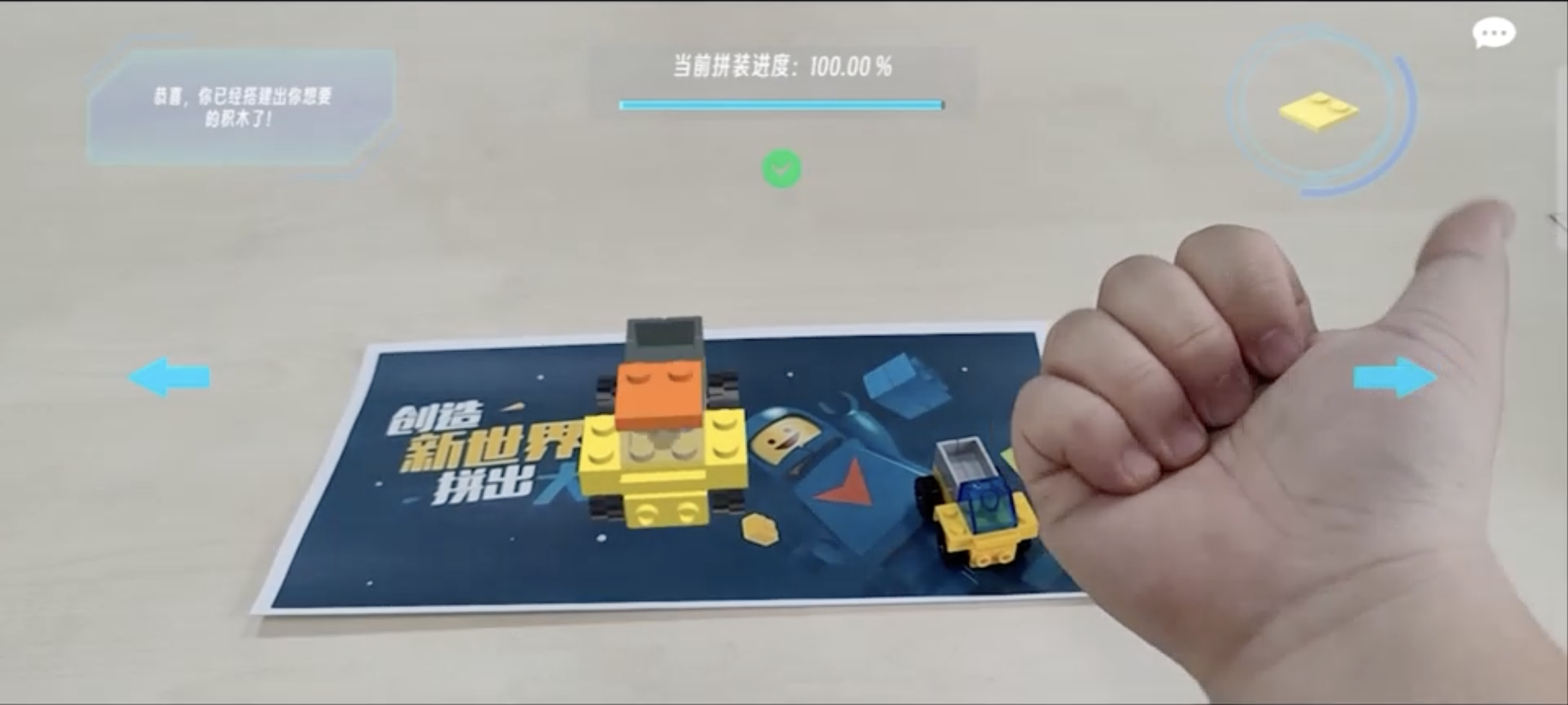}
     \caption{LEGO Brick Assembly. We illustrate several use cases in the demo of BrickDream.\footnotemark\ }
    \end{subfigure}
\caption{Examples of fine-grained assembly in \ac{MR} systems.}
\label{fig:demos}
\end{minipage}
% \vspace{-5mm}
\end{figure}
\footnotetext{\url{https://www.youtube.com/watch?v=KkZKL3aKMJs}}

\begin{figure*}[ht!]
    \centering
    \includegraphics[width=1\linewidth]{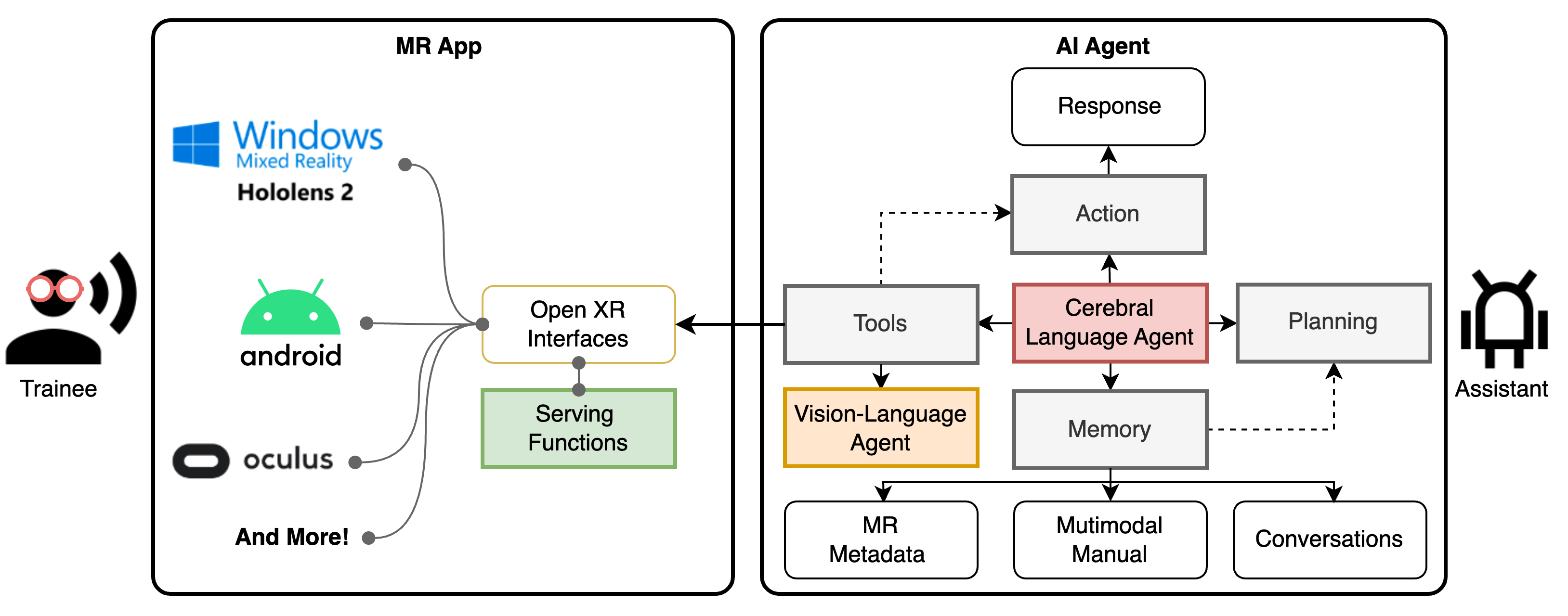}
    \caption{The proposed autonomous workflow, involving an \acs{AI} agent interacting with an \acs{MR} application. The \acs{AI} agent comprises a core cerebral language agent, which interacts with a vision-language agent to interpret the multimodal context into metadata, which can be utilized by the cerebral language agent iteratively. The \acs{MR} application interacts with AI agents by serving functions as external tools.}\label{fig:workflow}
    % \vspace{-5mm}
\end{figure*}

%% 1. Smarter assistant for Industry 4.0 and training assistance for industrial assembly 
The advent of ``Industry 4.0'', centered on the concept of smart manufacturing, presents a landscape with both opportunities and challenges for enhancing production efficiency~\cite{goel2020robotics,becue2021artificial,jan2023artificial}.
Training assistance for automating and accelerating industrial assembly is in huge demand across various manufacturing applications, such as furniture manufacturing~\cite{you2022human}, industrial product assembly~\cite{funk2017working}, and car assembly~\cite{bellalouna2020fiaar}.

%%% 2. VR/AR's assistance for industrial manufacturing assembly
\Acf{MR}, encompassing both \ac{VR} and \ac{AR}, spans a spectrum from fully real environments to ``matrix-like'' virtual environments, showing promise for industrial manufacturing assembly tasks~\cite{gavish2015evaluating,stender2021using,butaslac2022systematic}.
% \Ac{MR}, encompassing both \Ac{VR} and \ac{AR}, spans a spectrum from fully real environments to ``Matrix-like" virtual environments~\cite{skarbez2021revisiting}.
% \Ac{VR} and \ac{AR} systems are increasingly promising for industrial manufacturing assembly tasks~\cite{gavish2015evaluating,stender2021using,butaslac2022systematic}.
These multimodal, interactive, user-centric environments provide a solution for trainees who experience significant cognitive workload for training~\cite{hou2013study,botto2020augmented,dalim2020using}.
However, the assistance of a senior person as a trainer is required, either in person or remotely~\cite{fidalgo2023survey}.

%%% 3. AI integration with traditional models and challenges
To advance intelligent virtual assistants, traditional work leverages \ac{NLP} techniques~\cite{li2021bot,li2021can,li2022bringing, colabianchi2023human} and reinforcement learning~\cite{sloan2022adaptive} to promote human-machine interactions. Large language models (\acp{LLM}), as the new era of prevalent \ac{NLP} techniques, have been observed to elicit diverse interaction patterns across tasks, demonstrating their versatility and feasibility~\cite{mahmood2023llm}.
However, 
\begin{enumerate*}[label=(\roman*)]
    \item tailoring assistant services by grounding interactions; and 
    \item understanding users' situated multimodal contexts 
\end{enumerate*}
remain challenging and under-explored~\cite{dong2023towards}.

%%% AI agents
To this end, we introduce an autonomous workflow (see Figure~\ref{fig:workflow}) tailored for integrating \ac{AI} agents into \ac{MR} applications for fine-grained training. 
We present a demonstration of a multimodal fine-grained training assistant within a toy \ac{MR} application for LEGO brick assembly. 
Specifically, we design a cerebral language agent that integrates the \ac{LLM} with memory, planning, and interaction with \ac{MR} serving functional tools and a vision-language agent, enabling agents to decide their actions based on experiences.
Then, we introduce \OurData{}, a multimodal fine-grained assembly dataset synthesized automatically by a commercial \ac{LLM}. 
This dataset comprises 65 multimodal instruction manuals, 1,423 conversations with vision-language pairs, serving usages of 18 functional tools in an \ac{MR} environment. 
Additionally, several prevailing open-resource \acp{LLM} are presented as benchmarks, assessing their performance with and without fine-tuning on the proposed dataset.
Furthermore, we anticipate that the broader impact of this workflow will advance the development of smarter assistants for seamless user interaction in \ac{MR} environments, fostering research in both \ac{AI} and \acs{HCI} communities.
% Industry 4.0 refers to the integration of automation and data exchange in manufacturing. 
% Industry 5.0 is a new concept that focuses on collaboration between humans and machines.

% \acp{LLM} could enable virtual assistants to understand ambiguous instructions, break down complex goals into executable steps, and autonomously complete chained tasks to fulfill user requests [10].

We summarize our contributions as follows:
\begin{itemize}[leftmargin=*, nosep]
\item We design a workflow, which integrates autonomous \ac{AI} agents for fine-grained assembly assistance in an \ac{MR} demonstration. 
\item We create a multimodal manual-grounded fine-grained assembly conversation dataset in the \ac{MR} context.
\item We assess several open-resource \acp{LLM} as benchmarks, evaluating their performance with and without fine-tuning on the proposed dataset.
\end{itemize}

%% file: 2_related-work.tex
\section{Related Work}\label{sec:related-work}
% \begin{figure*}[ht!]
%     \centering
%     \includegraphics[width=1\linewidth]{figure/LEGO-ARTA-workflow.png}
%     \caption{The proposed autonomous workflow, involving an \acs{AI} agent interacting with a \acs{MR} application. The \acs{AI} agent comprises a core cerebral language agent, which interacts with a vision-language agent to interpret multimodal context into metadata, which can be utilized by the cerebral language agent iteratively. The \acs{MR} application seamlessly interacts with AI agents by serving functions as external tools.}\label{fig:workflow}
% \end{figure*}

We summarize previous research concerning multimodal datasets and virtual dialogue assistants within the realm of \ac{MR}.
% For thoroughness, we provide preliminaries concerning multimodal datasets and virtual dialogue assistants within the realm of \ac{MR}.
%We summarize previous research on multimodal datasets and virtual dialogue assistants in the context of \ac{MR}.

\input{table/datasets}

\subsection{Multimodal Datasets towards \ac{MR}}
% Most datasets focus on multimodal sensor data to study on the interactions in \ac{MR}.
Traditional multimodal datasets focus on the interactions with sensor data~\cite{jonell2018lrec} between human-human or human-robot, and only a few of them provide small-scale task-oriented dialogues, such as OFAI-MMTD~\cite{schreitter2016ofai} and ~\citet{kontogiorgos2018multimodal}, and Chinese Whispers~\cite{kontogiorgos2020lrec}.
ScanScribe~\cite{zhu2023vista} releases a 3D scene-text pairs dataset for 3D vision and text alignment learning.
HoloAssistant~\cite{wang2023holoassist} provides a dataset containing 350 unique instructor-performer pairs with \ac{AR} metadata to perceive, reason, and interact in the physical world.
However, the conversations are not publicly available.

Recent studies have concentrated on multimodal datasets with conversations.
MDC~\cite{narayan2019collaborative} presents a collection of 509 human-human conversations in the Minecraft \ac{VR} games.
CerealBar~\cite{suhr2019executing} creates 1,202 human-to-human conversations that map user instructions to system actions in a situated \ac{VR} game environment.
CVDN~\cite{thomason2020jointly} collects 2,050 human-robot conversations on Amazon Mechanical Turk for improving parsing and perception for natural language commands.
Teach~\cite{teach} builds over 3,000 human–human, interactive dialogues to complete household tasks in the simulation.

Different from those aforementioned datasets, \OurData{} gathers 1,423 synthetic natural conversations between trainers and trainees.
Unlike robotic commands, the length of utterances is relatively longer.
Furthermore, these conversations are generated by grounding both on an instruction manual and responses from an \ac{MR}, ensuring that the simulated conversations closely resemble natural human language.
We compare the statistics of the above datasets in Table~\ref{tab:dataset_comparision}.

\subsection{Virtual Dialogue Assistants for \ac{MR}}

%%% 传统方法：virtual dialogue assistants in VR/AR/MR 
Conventional efforts focus on creating virtual assistants for human-machine interactions using NLP techniques~\cite{li2021bot,li2021can,li2022bringing, colabianchi2023human} and reinforcement learning~\cite{sloan2022adaptive}.
%Traditional work aims to develop virtual assistants for human-machine interactions by \ac{NLP} techniques~\cite{li2021bot,li2021can,li2022bringing, colabianchi2023human} and reinforcement learning~\cite{sloan2022adaptive}.
%
%%% 新的方法用LLM：
\acp{LLM}, representing the forefront of contemporary NLP techniques, hold tremendous promise for advancing towards the next generation of intelligent assistants~\cite{naveed2023comprehensive}.
%\acp{LLM}, as the new era of prevailing \ac{NLP} techniques, have immense potential in advancing towards next-generation intelligent assistants~\cite{naveed2023comprehensive}.
%
%%% 最新的方法用LLM+automousmous agent
%Towards efficient generative large language model serving: A survey from algorithms to systems~\cite{miao2023towards}. 
The recent remarkable achievements of \acp{LLM} have spurred a growing interest in utilizing them to address a great variety of complex tasks~\cite{zhang2023dialogstudio}, with particular attention being drawn to LLM-augmented autonomous agents~\cite{yao2022react,huang2022language,shinn2023reflexion,madaan2023self}.

Autonomous agents expand the capabilities of LLMs into sequential action execution, demonstrating their proficiency in interacting with environments and addressing complex tasks through data collection~\cite{wang2023survey,liu2023bolaa}.
A crucial aspect of this advancement relies on the capacity of LLMs to generate and interpret images, enabling them to access visual content and provide inputs, thereby integrating with \ac{MR} environments~\cite{oyanagi2023virtual,wei2024editable}.
Regarding skill training, autonomous agents and LLMs can create immersive learning experiences that blend virtual and physical environments. For instance, students can utilize them to explore workflows and concepts in a more interactive and engaging manner~\cite{gong2023mindagent,li2024integrating}.
In the context of \ac{MR} serving as a sandbox~\cite{li2023metaagents} for LLMs and autonomous agents, the relationship is mutually beneficial. 
\ac{MR} offers a secure, adaptable, and regulated setting for training models~\cite{naihin2023testing}. 
\ac{LLM}-powered autonomous agents together with \ac{MR} hold the potential to revolutionize our interaction with the digital world~\cite{xu2023urban}.

%%% 总结对比我们的workflow
% Autonomous \ac{AI} agents~\cite{wang2023survey,guan2023intelligent,durante2024agent}, leading new era of , there is new potential to seamlessly integrate these agents into \ac{MR} environments and automatically discover business needs.
The convergence of \acp{LLM}, autonomous agents, and \ac{MR} presents both excitement and challenges. 
As \ac{MR} training experiences become more realistic and personalized, they demand larger amounts of data, encompassing detailed information about trainees' behaviors, preferences, and interactions. 
Ensuring the availability and reusability of this data poses a significant challenge.
Overall, our workflow's ultimate goal is to enhance \ac{MR} training experiences by facilitating more natural language interactions, generating precise 3D models of real-world objects~\cite{li2023m3dbench}, and fostering dynamic and interactive experiences. 
While challenges remain~\cite{xi2023rise,ayache2023extended}, the potential of this powerful technological fusion offers numerous exciting possibilities that could revolutionize personalization in virtual experiences. This entails the development of dedicated workflows and datasets.

%% file: table/datasets.tex
\begin{table*}[t!]
\small
\centering
\begin{tabular}{@{}llrrrrll@{}}
\toprule
          & Domain                             & \#Conv. & \#Utt.  & \#Token & \#AvgUtt. & \#AvgToken                 &  \\ \midrule
MDC       & Minecraft Building           & 509            & 15,926       & 113,116 & 30.7                   & 7.9 (Architect) / 2.9 (Builder)      &  \\
CerealBar & Instruction Following  & 1,202          & 23,979       & 3,641   & 19.9                   & 14.0 (Instructor) / 8.5 (Follower)   &  \\
CVDN      & Navigation & 2,050          & 12,361       & 2,223   & 6.0                    & 33.5 (Navigators) / 48.1 (Oracles) &  \\
TEACH     & Household                    & 3,215          & 45,000 & 3,429   & 13.7                   & 5.7 (Commander) / 3.8 (Follower)  &  \\ \midrule
LEGO-MRTA & LEGO Assembly                & 1,423          & 35,131       & 7,173   & 24.8                   &  26.6 (Trainer) / 12.7 (Trainee)    &  \\ \bottomrule
\end{tabular}
\caption{Comparison of dialogue datasets towards \ac{MR}.}
\label{tab:dataset_comparision}
% \vspace{-5mm}
\end{table*}

%% file: 3_workflow.tex
\section{Fine-Grained Training Workflow}\label{sec:workflow}
% \begin{figure*}[ht!]
%     \centering
%     \includegraphics[width=1\linewidth]{figure/LEGO-ARTA-workflow.png}
%     \caption{The proposed autonomous workflow, involving an \acs{AI} agent interacting with a \acs{MR} application. The \acs{AI} agent comprises a core cerebral language agent, which interacts with a vision-language agent to interpret multimodal context into metadata, which can be utilized by the cerebral language agent iteratively. The \acs{MR} application seamlessly interacts with AI agents by serving functions as external tools.}\label{fig:workflow}
% \end{figure*}

In this section, we describe the proposed workflow (See Figure~\ref{fig:workflow}) that advances \ac{AI} agents towards \acs{MR} guided fine-grained training.

\subsection{Definition of Fine-Grained Training}\label{sec:task}
In the context of fine-grained training, we anticipate the ability to 
\begin{enumerate*}[label=(\roman*)]
    \item accurately follow professional training instructions documented in an instruction manual; and 
    \item be sensitive to detailed visual information, ultimately for complex industrial assembly tasks, as illustrated in Figure~\ref{fig:demos} (a).
\end{enumerate*}

We define the following two roles during a training session:
\begin{itemize}[leftmargin=*,nosep]
\item \textbf{User:} A human trainee who aims to acquire expertise and will work on fine-grained assembly tasks through interaction with the \ac{MR} environment.
\item \textbf{Assistant:} A virtual AI agent who will be able to assist the trainees in training and
respond to their inquiries.
It offers support with
\begin{enumerate*}[nosep, label=(\roman*)]
\item a conversation agent that replies to trainees' requests and provides guidance grounded in the instruction manual;
\item an interface for users to interact with the \ac{MR} environment; and
\item a vision-language agent that understands and transmits users' visual context to language.
\end{enumerate*}
\end{itemize}

\subsection{Autonomous \ac{AI} Agent}
We design the autonomous \ac{AI} agent with a chain of two agents, namely 
\begin{enumerate*}[nosep, label=(\roman*)]
\item a cerebral language agent that serves to reply to trainees' requests, provide guidance, interact with \ac{MR} and the vision-language agent; and 
\item a vision-language agent that understands and transmits users' visual context to language, which is then utilized by the cerebral language agent for planning.
\end{enumerate*}

\subsubsection{Cerebral Language Agent}\label{sec:llm_agent}
% center area: training knowledge (general and LEGO-specific)
Inspired by the concept of \ac{LLM}-powered autonomous agents~\cite{wang2023survey}, we develop a cerebral language agent that incorporates an \ac{LLM} with \textit{memory}, \textit{planning}, and functional \textit{tools} that can interact with \ac{MR} application, thereby enabling agents to make decisions regarding their \textit{actions} based on past experiences.
It can handle multimodal inputs, such as instruction manuals, historical conversations, and metadata within \ac{MR} environments, and subsequently generate actions (i.e., responses or API calls for the \ac{MR} application).
The scope of responsibility of the agent is defined in a system prompt (See P2, Table~\ref{tab:template_conversation_generation}, Appendix~\ref{sec:appendix})
Notably, it is able to alleviate the challenges (See \S\ref{sec:introduction}): 
\begin{enumerate*}[nosep, label=(\roman*)]
\item it tailors assistant services by seamlessly interaction with \ac{MR} applications to discover the business needs gradually;
\item it interacts with a vision-language agent (See \S\ref{sec:vlm_agent}), which facilitates the capability of understanding the multimodal context in \ac{MR} environments.
\end{enumerate*}

% and a \textbf{vision-language agent}. This 

\subsubsection{Vision-Language Agent}\label{sec:vlm_agent}
% visual speech area: scene description
The vision-language agent's mission is to bridge the gap between understanding visual context and language, enabling effective utilization by the cerebral language agent (See \S\ref{sec:llm_agent}) to conduct comprehensive planning for global optimization.
Its core is the \ac{VLM}, which is a task-driven large model that transmits vision input into language output needed by specific tasks.

In the context of LEGO assembly training, we observe two distinct patterns in LEGO instruction manuals (See an example in Figure~\ref{fig:example_manual}) and define the following two tasks:
\begin{enumerate*}[nosep, leftmargin=*, label=(\textbf{T\arabic*)}]
\item \textbf{Object detection.} 
Given an image or a sequence of images as input, the objective is to predict the position of an object requested in a query and generate output in the format of ``<Object> <Xleft> <Ytop> <Xright> <Ybottom>''.
For example, during assembly step 2, the \ac{AI} trainer might direct the trainee, ``Please gather the earth blue pair of legs and the silver metallic upper part of the body.'' In response, the trainee may ask, ``Is this the one?'' The vision-language agent is tasked with recognizing the object the trainee is referring to.
\item \textbf{Assembly state detection.} 
Given an image or a sequence of images as input, the objective is to identify if the current assembly state matches the reference state provided in the instruction manual.
For example, during assembly step 3, the vision-language agent is responsible for assisting the user's request, such as ``Am I assembling them correctly?''
\end{enumerate*}

% \subsubsection{Motor language agent}
% motor speech area: navigation

% \subsubsection{Auditory-speech agent}
% % auditory speech area: ASR and TTS

\subsection{Pilot \acs{MR} Application Design}\label{sec:xr_app}
We design an \ac{MR} application as a pilot to show intuitive demonstration.
First, we utilized a commercial \ac{LLM} to generate candidate user requirements using the prompt (See P1, Table~\ref{tab:template_conversation_generation}, Appendix~\ref{sec:appendix}) as input. 
Then, we brainstormed and discussed the generated user requirements within a group of researchers and developers and finalized 7 user requirements (See Table~\ref{tab:user_rerquirements}, Appendix~\ref{sec:appendix}) and 18 serving functional tools (See Table~\ref{tab:tool_description}).
We develop a standard \acp{API} to enable seamless interactions between functions in the \ac{MR} application and \ac{AI} agents.
% The information inflow is illustrated in (Figure~\ref{fig:information_flow}, Appendix~\ref{sec:appendix})
\input{table/tool_descriptions.tex}

%% file: table/tool_descriptions.tex
\begin{table*}[h!]
\centering
\small
\renewcommand{\arraystretch}{1.15} % Default value: 1
\begin{tabular}{lp{11cm}}
\hline
Tool Name & Description \\
\hline
StartAssemble & Initiate the assembly process. \\
NextStep & Move to the next assembly step. \\
FrontStep & Go back to the previous assembly step. \\
Explode & Trigger an explosion for detailed viewing. \\
Recover & Restore the initial state of AR objects after explosion. \\
FinishedVideo & End the assembly process and show a video of the assembled LEGO bricks. \\
ReShow & Repeat the current assembly step. \\
Enlarge & Enlarge or zoom out the current object. \\
Shrink & Shrink or zoom in the current object. \\
GoToStep & Go to the given assembly step number. \\
Rotate & Rotate the current object to a direction (``Up'', ``Down'', ``Left'', ``Right'', ``None''). \\
ShowPieces & Show all candidate LEGO pieces to be assembled. \\
HighlightCorrectComponents & Highlight correct attachment points and components. \\
GetCurrentStep & Get the number of the current step. \\
GetRemainingStep & Get the number of the remaining steps. \\
CheckStepStatusVR & Check whether the current step in Unity is accomplished correctly or not. 
% If the current assembly sequence recorded in Unity is the same as the manual assembly sequence, then it is correct, otherwise, it is incorrect. 
\\
APICallObjectRecognitionAR & Call the VLM agent to identify LEGO pieces based on the provided video streaming data from AR glasses and highlight the recognized pieces in the AR environment. \\
APICallCheckStepStatusAR & Call the VLM agent to determine whether the current assembly step is completed correctly or not, using the provided video streaming data from AR glasses as input. \\
\hline
\end{tabular}
\caption{Descriptions of serving tools in the pilot \ac{XR} application.}
\label{tab:tool_description}
% \vspace{-5mm}
\end{table*}

%% file: 4_dataset.tex
\section{Dataset Creation}\label{sec:dataset_creation}
In this section, we introduce how to use the proposed workflow (See \S\ref{sec:workflow}) to create a multimodal dialogue dataset in the \ac{MR} environment.
\subsection{Instruction Manual Crawling}
\begin{figure}[t!]
    \centering
    \includegraphics[width=1\linewidth]{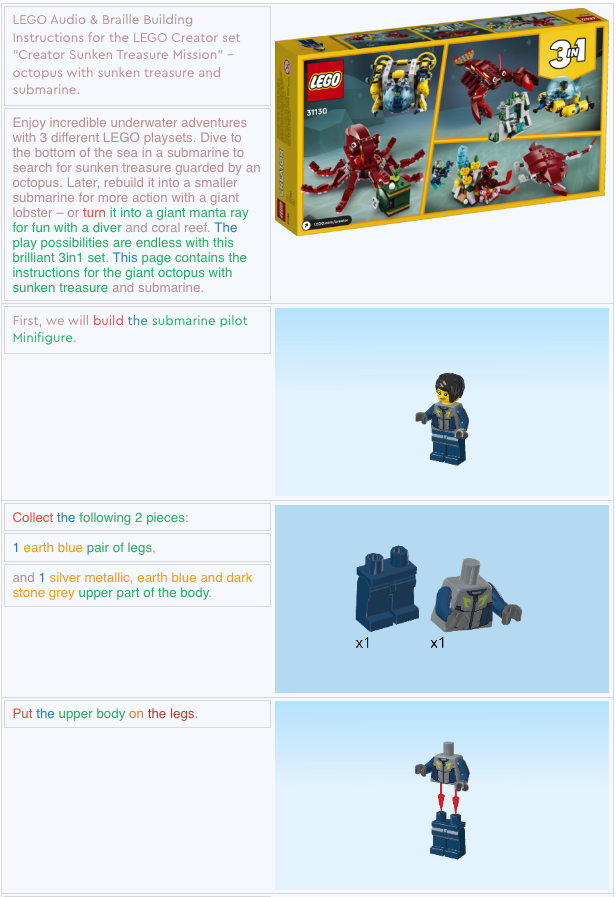}
    \caption{An example of LEGO instruction manual. It consists of a summary section at the beginning followed by three sequential instruction steps. Each step includes textual instructions paired with corresponding images to guide the assembly process.}
    \label{fig:example_manual}
\end{figure}

We crawled 65 multimodal instruction manuals for fine-grained training from the LEGO official website\footnote{\url{https://legoaudioinstructions.com/instructions}}.
A manual provides illustrated images and textual instructions on how to use, operate, assemble, and install a LEGO brick set.
The key sections of an instruction manual include: 
\begin{enumerate*}[label=(\roman*)]
\item a summary that describes the general information, such as topics and candidate parts for assembly. It is followed by
\item a sequence of multimodal step instructions. 
Each step contains a set of textual instructions and an illustration by image.
Key functional phrases such as theme entities are highlighted in textual instructions.
Here we show an example of a LEGO instruction manual in Figure~\ref{fig:example_manual}.
\end{enumerate*}

\subsection{Tool Response Generation}
First, we use the crawled instruction manuals and the well-designed prompt template to produce prompts as the \ac{LLM} input to generate user functional requirements and decide the serving functional tools.
Then, we randomly choose up to 6 tools for each conversation session and record the simulated responses generated from templates.

\subsection{\acs{VLM}-Based QA Construction}
First, we use the step instruction to construct a query, containing a special token (``[detection]'') for the object detection task and a single instruction in a step. 
Second, we employ a query and the aligned image as inputs for MiniGPT-v2~\cite{chen2023minigptv2,zhu2023minigpt}, generating inference output as an answer of the query in the format of ``<Object> <Xleft> <Ytop> <Xright> <Ybottom>''.
Last, we iterate through all instruction steps in a conversation session, repeating the above two steps to construct \ac{VQA} pairs.

\subsection{Multimodal Context-Aware Conversation Generation}
We generate conversations grounded on both the instruction manual and simulated tool responses using a commercial \ac{LLM}.
First, we reconstruct full instruction manuals with a summary and 10 step instructions because the average number of steps per manual is 215.3, which is quite long. 
This may limit the input tokens of an LLM and potentially distract the LLM with less grounding capability.
Second, we instantiate the designed prompt template (See P3, Table~\ref{tab:template_conversation_generation}, Appendix~\ref{sec:appendix}) with the chunked instruction manuals.
Last, we utilize a commercial LLM as the core of the proposed workflow to generate the conversations.
Specifically, the system prompt informs the language agent about its responsibilities.
The query prompt is used for each round of requests to generate a conversation.
The historical rounds of requests are tracked by memory.

\subsection{Dataset Statistics}
\input{table/dataset_instruction_manual}

We report the statistics of instruction manuals (See Table~\ref{tab:statistic_manual}) and conversations (See Table~\ref{tab:dataset_comparision}).

We obtain 65 instruction manuals as grounding to lead a commercial \ac{LLM} to generate 1,423 human-human natural conversations between trainers and trainees.
Each instruction manual can make 21.9 conversations on average. 
% {\color{red}{make what? there should be an object. besides, 1423/65 = 21.89}}
Theoretically, the amount of conversations can be enlarged by multiple times of requests.
However, we focus on showcasing how to create meaningful datasets automatically.
%%%
We construct 26,405 context-response pairs from generated conversations and \ac{VQA} pairs as data samples.
The average length is 107 tokens for the context and 145 tokens for the response utterance.
We utilize 21.10k samples for fine-tuning open-resource \acp{LLM} to enhance the instruction-following capability and evaluate their performance on 5.25k test samples~\footnote{\url{https://huggingface.co/datasets/voxreality/vox_arta_lego_v2}}.
%%%
Compared with existing datasets, \OurData{} ensures that the simulated conversations closely resemble natural human language due to the design of the simulation method.

%% file: table/dataset_instruction_manual.tex
\begin{table}[t!]
\footnotesize
\centering
\setlength{\tabcolsep}{25pt}
\resizebox{\linewidth}{!}{ 
\begin{tabular}{@{}lr@{}}
\toprule
LEGO-MRTA Instruction Manual            &            \\ 
\midrule
\#Manual  & 65        \\  
\#InstructionStep & 13,994 \\
\#Token & 8,676\\
\#Theme Entity  & 2,412 \\
\#AvgInstructionStep      & 215.3                     \\ 
\#AvgConversation      & 21.9                     \\ 
% \#Tool  & 18         \\  \midrule
Modalities          & Text, Image \\ 
% \acsp{VQA}            & xxx     \\ 
\bottomrule
\end{tabular}
}
\caption{Statistics of instruction manuals in the \OurData{} dataset.}\label{tab:statistic_manual}
\end{table}
% \vspace{-6mm}

%% file: 5_experimental_setup.tex
\section{Experimental Setup}
\input{table/main_results}
\subsection{\ac{LLM} Benchmarks}
We consider several prevailing 7B open-source decoder-only \acp{LLM} as benchmarks, considering privacy concerns associated with fine-grained training in manufacturing.
\begin{itemize}[nosep, leftmargin=*]
    \item \textbf{BLOOM}~\cite{le2211bloom} is pretrained on the multilingual ROOTS corpus, offering multilingual capabilities for various \ac{NLP} tasks.
    \item \textbf{Falcon-instruct}~\cite{falcon40b} is pretrained on a large corpus of RefinedWeb data and fine-tuned on mixed chat and instruct datasets.
    % under the permissive Apache 2.0 license.
    \item \textbf{Llama2-Chat}~\cite{touvron2023llama} is a pretrained and fine-tuned generative text model optimized specifically for dialogue tasks, ensuring high-quality conversational responses.
    \item \textbf{Vicuna1.5}~\cite{zheng2023judging} is a chat assistant derived by fine-tuning Llama 2 on user-shared conversations collected from ShareGPT.
    \item \textbf{OpenChat3.5}~\cite{wang2023openchat} is a chat model fine-tuned with the C-RLFT strategy on mixed-quality data, achieving performance comparable to larger models like ChatGPT.
    \item \textbf{XVERSE}\footnote{\url{https://github.com/xverse-ai/XVERSE-7B/blob/main/README_EN.md}} is a versatile model supporting 8k context length, ideal for longer multi-round dialogues, knowledge question-answering, and summarization tasks, trained on a diverse dataset of 2.6 trillion tokens.
    \item \textbf{BlueLM-Chat}\footnote{\url{https://huggingface.co/vivo-ai/BlueLM-7B-Chat}} is a large-scale language model optimized for chat tasks, offering improved context understanding. 
    % under the permissive Apache 2.0 license.
    \item \textbf{Qwen-Chat}~\cite{bai2023qwen} is a chat model that fine-tunes the pretrained Qwen model using human alignment techniques.
    \item \textbf{Mistral-Instruct}~\cite{jiang2023mistral} is a fine-tuned version of the Mistral-7B-v0.1, specifically tailored for instruction-based tasks using publicly available conversation datasets.
\end{itemize}

% \vspace{-3mm}
\subsection{Evaluation Metrics}
We evaluate the performance in terms of both overlap (BLUE-n, ROUGE-n) and informativeness (ToolACC, ThemeACC):
\begin{itemize}[nosep, leftmargin=*]
    \item \textbf{BLUE-n} measures precision, which measures the ratio of n-grams in the generated responses that match those in the reference responses. We consider $n=4$.
    \item \textbf{ROUGE-n} measures recall, which calculates the ratio of n-grams in the reference responses that are captured by the generated responses. Here we consider $n=1,2,L$ and $L$ denotes the number of longest common subsequences.
    \item \textbf{ToolACC} is defined as the ratio of correctly mentioned entities by the generated responses, compared to the reference response, from a list of serving tools.
    \item \textbf{ThemeACC} is defined as the ratio of correctly mentioned entities compared to the reference response, from a list of theme entities obtained from the instruction manual.
\end{itemize}

% \subsection{User experiments}
\subsection{Implementation Details}
The implementation of the workflow is based on LangChain.\footnote{\url{https://python.langchain.com/docs/get_started/introduction}}
The model ``gpt-3.5-turbo-16k-0613'' is used as the commercial \ac{LLM} for generating data, e.g., conversations, user requirements, and serving functions.
The MiniGPT4-v2\footnote{\url{https://github.com/Vision-CAIR/MiniGPT-4}} is used as the \ac{VLM} to detect the object, followed by simple rules to generate vision-language pairs.

We employ \ac{LoRA} to conduct parameter-efficient fine-tuning of 7B open-source \acp{LLM} on the proposed dataset using the framework proposed by \citet{llama-factory}.
Specifically, the maximum sequence length is set to 1024 and the learning rate is 5e-05. 
The model is trained for 3 epochs with a per-device batch size of 4, and accumulated gradients every 4 steps.
A cosine learning rate scheduler is employed, with a maximum gradient norm of 1.0. 
We log results every 5 steps and save model checkpoints every 100 steps. 
Warm-up steps are set to 0. 
\ac{LoRA} is used with a rank of 8 and a dropout rate of 0.1 for regularization.
All experiments are run on NVIDIA A100 SXM4 40GB GPUs. 
% The demonstration is deployed on two devices, i.e., HoloLens 2 and Android. 

%% file: table/main_results.tex
% \usepackage{booktabs}
% \usepackage{multirow}
\begin{table*}[t!]
\resizebox{\linewidth}{!}{ 
\centering
\setlength\tabcolsep{2pt}
% \small
\begin{tabular}{@{}lrrrrrrrrrrrrrrrrrrrrrrrr@{}}
\toprule
Model & \multicolumn{4}{c}{BLEU-4} & \multicolumn{4}{c}{ROUGE-1} & \multicolumn{4}{c}{ROUGE-2} & \multicolumn{4}{c}{ROUGE-L} & \multicolumn{4}{c}{ToolACC (\%)} & \multicolumn{4}{c}{ThemeACC (\%)} \\ 
\cmidrule(lr){2-5} \cmidrule(lr){6-9} \cmidrule(lr){10-13} \cmidrule(lr){14-17} \cmidrule(lr){18-21} \cmidrule(lr){22-25}
PEFT (LoRA)                       &     & /wo    & /w    &    &     & /wo    & /w    &     &     & /wo    & /w    &     &     & /wo    & /w    &     &     & /wo    & /w    &     &      & /wo    & /w    &     \\ 
% \cmidrule(r){1-1}
\midrule
BLOOM      &  & 2.88   & 54.07 &  &  & 20.49  & 61.91 &  &  & 6.50   & 49.52 &  &  & 3.78   & 58.63 &  &  & 49.62  & 77.86 &  &  & 26.30  & 64.61 &  \\
Falcon      &  & 5.38   & 10.30 &  &  & 8.68   & 11.33 &  &  & 4.25   & 7.41  &  &  & 5.11   & 10.20 &  &  & 22.79  & \underline{17.65} &  &  & 12.66  & \underline{10.81} &  \\
Llama2-Chat             &     & 10.23     & 30.53    &    &     & 18.59     & 40.65    &     &     & 7.41      & 25.48    &     &     & 10.59     & 32.91    &     &     & 21.37     & 55.73    &     &      & 47.20     & 55.51    &     \\
Vicuna1.5           &     & 14.11     & 54.71    &    &     & 29.30     & 62.64    &     &     & 14.21     & 50.47    &     &     & 15.48     & 59.36    &     &     & \textbf{52.67}     & 78.12    &     &      & \textbf{69.69}     & \underline{66.79}    &     \\
OpenChat3.5         &     & 22.00     & \underline{6.94}     &    &     & 29.70     & 34.51    &     &     & 15.69     & 23.69    &     &     & 22.50     & 11.36    &     &     & 51.97     & 74.02    &     &      & 58.19     & \textbf{81.90}    &     \\
XVERSE      &     & 22.42     & 53.55    &    &     & 28.45     & 61.54    &     &     & 14.31     & 49.77    &     &     & 22.39     & 58.03    &     &     & 49.62     & \textbf{83.97}    &     &      & 57.53     & 71.10    &     \\
BlueLM      &  & 22.72  & 55.69 &  &  & 30.40  & 63.52 &  &  & 14.98  & 51.58 &  &  & 23.76  & 60.35 &  &  & 48.15  & 82.22 &  &  & 47.51  & 68.08 &  \\ 
Qwen   &  & 24.82  & \textbf{59.78} &  &  & 31.44  & \textbf{66.95} &  &  & 17.69  & \textbf{55.95} &  &  & 25.66  & \textbf{64.26} &  &  & 45.71  & 77.14 &  &  & 54.96  & 71.17 &  \\
Mistral     &     & \textbf{25.87}     & 54.17    &    &     & \textbf{33.32}     & 62.07    &     &     & \textbf{17.99}     & 49.40    &     &     & \textbf{26.32}     & 58.62    &     &     & 49.62     & 78.20    &     &      & 54.80     & 66.65    &     \\ 
\bottomrule
\end{tabular}
}
% \vspace{-3mm}
\caption{Benchmarking \acp{LLM} on our \OurData{} dataset, without (/wo) and with (/w) \ac{PEFT} using \ac{LoRA}. 
The bold font indicates the highest score in each column. 
The underline indicates the performance decreases after fine-tuning. 
% {\color{red}{is decrease a noun or a verb here? I guess we should use decreases and add that after indicates}}
}\label{tab:benchmarks}
% \vspace{-5mm}
\end{table*}

%% file: 6_outcomes.tex
\section{Outcomes}

\begin{figure*}[t!]
    \centering
    \includegraphics[width=1\linewidth]{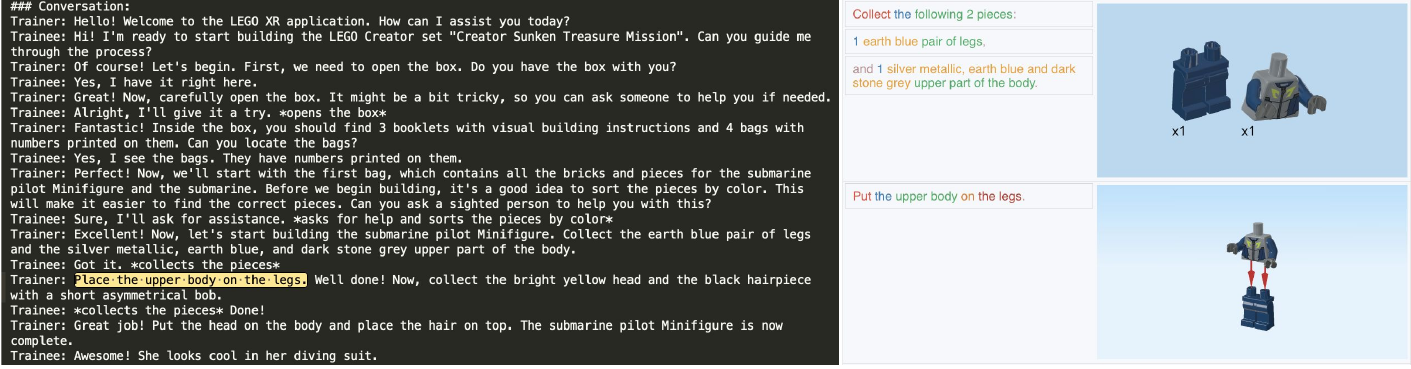}
    \vspace{-6mm}
    \caption{An example of the generated conversation (left) and the grounding step instructions (right).}
    \label{fig:example_conversation}
    \vspace{-5mm}
\end{figure*}
\begin{figure}[t!]
    \centering
    \includegraphics[width=1\linewidth]{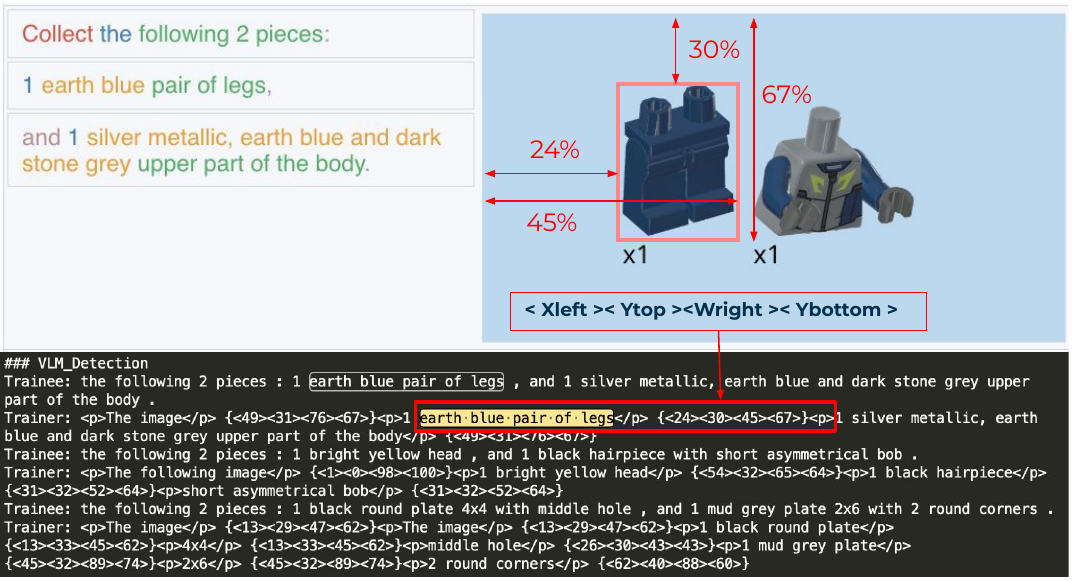}
    \vspace{-6mm}
    \caption{An example of vision-language pair (lower) and the grounding step instructions (upper).}
    \label{fig:example_vqa}
    \vspace{-5mm}
\end{figure}

\subsection{Evaluation on Benchmark \acp{LLM}}
Table~\ref{tab:benchmarks} shows the performance on 9 prevailing open-source \acp{LLM}, without and with fine-tuning on the \OurData{} dataset.

First, after fine-tuning, the performance of all models gets dramatically improved in terms of all metrics, except for the results that are underlined.
This demonstrates the feasibility and effectiveness of tailoring \acp{LLM} for fine-grained training in \ac{MR} environments.
In addition, this shows the proposed \OurData{} dataset contains distinct characteristics that have not been captured by existing publicly available datasets.

Second, there exists a trade-off between overlap and informativeness evaluation.
For example, for metrics concerning overlap, Qwen consistently achieves the highest scores; while its informativeness is inferior to best models, i.e., ToolACC is 6.83\% lower than that of XVERSE, ThemeACC is 10.73\% lower than that of OpenChat3.5.

Third, the choice of backbone \acp{LLM} inherently impacts the performance of both overlap and informativeness.
We compute the standard deviation for each metric over all models to gauge the variability in performance scores: BLEU-4 (19.60), ROUGE-1 (17.79), ROUGE-2 (16.02), ROUGE-L (20.64), ToolACC (19.85), and ThemeACC (19.19).
% Another noteworthy consideration is the licensing terms. 
% For example, Falcon-instruct and BlueLM-Chat are both offered under the permissive Apache 2.0 license, permitting commercial utilization without any royalties or restrictions.

\subsection{Case Study}

Figure~\ref{fig:example_conversation} shows an example to intuitively verify the feasibility of generating conversation based on an instruction manual.
As highlighted in the trainer's utterance, ``Place the upper body on the legs'', this accurately conveys the instruction from the manual in a human-like manner.
The generated conversation is feasible at instruction-following capability.

Figure~\ref{fig:example_vqa} illustrates an example demonstrating the feasibility of constructing queries based on instructions from a manual to accurately request positions in a multimodal context.
We transferred the generated position and highlighted a frame with markers.
We observed that the prediction was relatively accurate.
Additionally, another aspect we observe that needs improvement in the future is the redundant output from the \ac{VLM}.

%% file: 7_discussion.tex
\section{Discussion of Broader Impact}
The research presented in this paper offers a fully new environment to advance how workers are trained and get help by using \ac{MR} technologies.
By integrating AI assistants into MR environments, workers can tackle complex tasks more effectively. This innovation not only enhances worker productivity but also reduces training costs for companies, as it eliminates the need for expert instructors to be physically present for employee training sessions.

%% file: 8_conclusion.tex
\section{Conclusion}
In this work, we introduce an autonomous workflow to develop smarter multimodal fine-grained training assistants in \ac{MR} environments.
Specifically, we have designed a cerebral language agent that integrates \ac{LLM} with memory, planning, and interaction with MR tools, along with a vision-language agent. 
This integration enables agents to make decisions based on past experiences, thereby addressing the challenge of tailoring assistant services through grounded interactions.
We have designed a vision-language agent to better understand users’ situated multimodal contexts.
Notably, we have created a dataset for fine-grained training in \ac{MR}.
We have compared the performance of open-resource \acp{LLM} before and after fine-tuning using this dataset.
We aim to facilitate the development of smarter assistants for seamless user interaction, fostering research in \ac{AI} and \acs{HCI} communities.

%% file: _ack.tex
\section*{Acknowledgements}
This work was supported through the Horizon Europe research and innovation programme, under grant agreement No 101070521 (VOXReality). 
This work used the Dutch national e-infrastructure with the support of the SURF cooperative grant (No. EINF-7705).
Views and opinions expressed are however those of the author(s) only and do not necessarily reflect those of the European Union or Directorate-General for Communications Networks, Content and Technology. Neither the European Union nor the granting authority can be held responsible for them.

%% file: _appendix.tex
\input{table/prompt_template_and_task_instruction}
\input{table/user_requirements.tex}
\section{Appendix}\label{sec:appendix}
%%%%%%%%%%%%%%%%%%%%%%%%%%%%%%%%%%%%%%%%%%%%%%%%%%%%%%%%%%%%%%%%%%%%%%%% 1. Prompt templates
\subsection{Prompt Templates}
% \label{sec:appendix_prompt}

%%%%%%%%%%%%%%%%%%%%%%%%%%%%%%%%%%%%%%%%%%%%%%%%%%%%%%%%%%%%%%%%%%%%%%%% 2. Generated user requirements
\subsection{Generated User Requirements}
% \label{sec:appendix_user_requirement}

%%%%%%%%%%%%%%%%%%%%%%%%%%%%%%%%%%%%%%%%%%%%%%%%%%%%%%%%%%%%%%%%%%%%%%%% 3. Analytical study of the dataset
\subsection{Qualitative Analysis of the Dataset}
%%%%%%
\begin{figure*}[t]
\centering
    \begin{subfigure}[b]{0.23\textwidth}
     \centering
     \includegraphics[width=\textwidth, trim={0 0 0 0},clip]{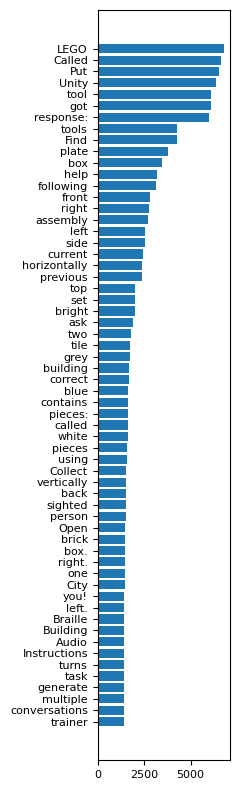}
     \caption{Instruction tokens.}
     \label{fig:distribution_manual}
    \end{subfigure}
    \hfill
    \begin{subfigure}[b]{0.278\textwidth}
     \centering
     \includegraphics[width=\textwidth]{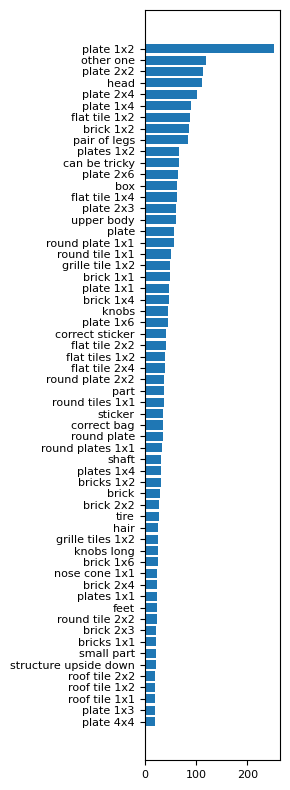}
     \caption{Theme entities.}
     \label{fig:distribution_entity}
    \end{subfigure}
    \hfill
    \begin{subfigure}[b]{0.22\textwidth}
     \centering
     \includegraphics[width=\textwidth]{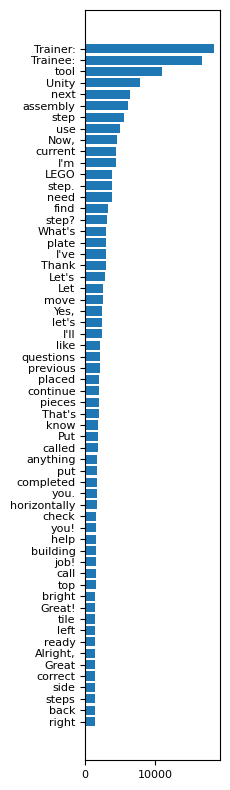}
     \caption{Conversation tokens.}
     \label{fig:distribution_conversation}
    \end{subfigure}
    \hfill
    \begin{subfigure}[b]{0.22\textwidth}
     \centering
     \includegraphics[width=\textwidth]{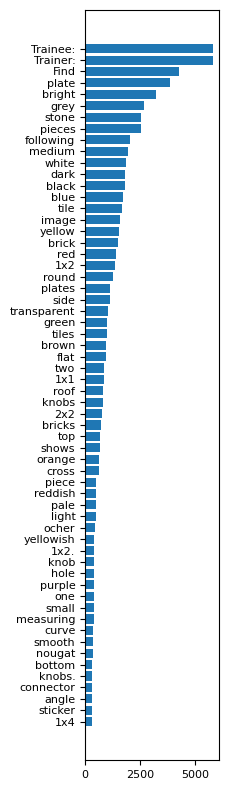}
     \caption{Vision-language tokens.}
     \label{fig:distribution_vlm}
    \end{subfigure}
\caption{Distribution of top 60 frequent tokens in the above four parts: (a) instructions, (b) entities in the manual, (c) conversations, and (d) vision-language pairs. The x-axis denotes frequency and the y-axis denotes tokens in four parts of the \OurData{} dataset.}
\label{fig:all-four-img-tokken}
\end{figure*}

The collective results as seen in the figure \ref{fig:all-four-img-tokken} illustrates that our simulated model has practical assistance capabilities as summarized below: 

% First, its 
\paragraph{Realistic simulation.}
LEGO is a well-known block building concept. The dataset simulates various real-world scenarios encountered during LEGO assembly tasks. By replicating factors such as piece variability, environmental conditions, and assembly constraints, the dataset provides a realistic training environment for machine learning models. This realism enhances the model's ability to generalize to unseen situations, ensuring reliable performance in diverse assembly settings.

\paragraph{Diversity in task difficulty.}
From simple structures to intricate designs, the dataset exposes the model to diverse assembly scenarios, enabling it to learn robust representations of LEGO building principles. This diversity fosters adaptability in the model, empowering it to tackle simple to difficult or probably novel assembly tasks with confidence and efficiency.

\paragraph{Transfer learning to other tasks.}
The dataset is structured to facilitate transfer learning, allowing knowledge and representations learned from one assembly task to be applied to related tasks or domains. By leveraging pre-trained models or features learned from similar assembly tasks, machine learning models can bootstrap their learning process on new assembly tasks. This transfer learning capability accelerates model adaptation to new environments and tasks, reducing the need for extensive retraining and improving overall training efficiency.

\subsubsection{Instruction Tokens (Figure \ref{fig:distribution_manual})}

Analyzing the provided instruction tokens, we can derive several factors that contribute to the usability and effectiveness of our simulated dataset for \ac{XR} training and assembly training:

\begin{itemize}[nosep, leftmargin=*]
    \item \textbf{Clear instructional guidance.} 
    Tokens like ``Put,'' ``Find,'' ``Collect,'' and ``Open'' provide clear and concise instructions for performing various assembly tasks. These instructions guide users by the assembly process step-by-step, ensuring clarity and direction in the training environment.
    \item \textbf{Spatial orientation and manipulation.}
    Tokens such as ``front,'' ``right,'' ``left,'' ``horizontally,'' and ``vertically'' offer spatial orientation cues, helping users understand the spatial relationships between LEGO components and how to manipulate them during assembly. This spatial awareness enhances users' ability to accurately position and align LEGO pieces.
    \item \textbf{Feedback and assistance.} 
    Tokens like ``help'' and ``response'' indicate provisions for feedback and assistance within the training environment. Offering assistance and feedback helps users troubleshoot issues, learn from mistakes, and improve their assembly skills over time, enhancing the learning experience.
    \item \textbf{Multimodal learning}: The inclusion of tokens like ``Audio Instructions'' suggests the incorporation of multimodal learning techniques within the training environment. Integrating audio instructions alongside visual cues enhances usability by catering to different learning styles and preferences, making the training experience more accessible and engaging for users.
    \item \textbf{Adaptive learning.}
    Tokens such as ``current'' and ``previous'' imply a dynamic learning environment where users can track their progress and revisit previous steps if needed. Adaptive learning features enhance usability by allowing users to learn at their own pace, review concepts as needed, and progress through the training material in a structured manner.
    \item \textbf{Interactive learning environment.}: The presence of tokens like ``conversations'' and ``trainer'' indicates an interactive learning environment where users can engage in dialogue and receive guidance from trainers or virtual assistants. Interactivity enhances usability by promoting engagement, collaboration, and active participation in the learning process, leading to more effective skill acquisition and retention.
\end{itemize}
The instruction tokens in our simulated dataset indicate clear guidance, and spatial orientation cues within an interactive learning environment.

\subsubsection{Theme Entities (Figure~\ref{fig:distribution_entity})}

Based on the theme entities provided, we analyze their relevance to the learning process:

\begin{itemize}[nosep, leftmargin=*]
    \item \textbf{Part identification.}
    Tokens such as ``plate 1x2,'' ``plate 2x4,'' and ``brick 1x1'' provide specific identifiers for different LEGO parts commonly used in assembly tasks. By including a variety of part identifiers, the dataset facilitates part recognition and identification, enabling the model to learn the characteristics and properties of each component.
    \item \textbf{Spatial orientation and configuration.}
    Tokens like ``head,'' ``upper body,'' and ``feet'' suggest the inclusion of assembly instructions related to spatial orientation and configuration of LEGO structures. Understanding the spatial arrangement of components is essential for accurate assembly, and these tokens help the model grasp the hierarchical structure of assemblies and the placement of parts within them.
    \item \textbf{Assembly techniques.}
    Tokens such as ``can be tricky'' and ``structure upside down`` hint at the inclusion of assembly techniques and strategies within the dataset. Learning various assembly techniques is crucial for efficiently building complex structures, and these tokens provide guidance on overcoming challenges and optimizing assembly processes.
    \item \textbf{Component variations.}
    Tokens like ``round plate 1x1,'' ``flat tile 2x4,'' and ``grille tile 1x2`` introduce variations of standard LEGO components, reflecting the diversity of parts encountered in real-world assembly scenarios. By including a range of component variations, the dataset contains different types of parts and adapts to varying assembly requirements.
    \item \textbf{Accessory identification.}
    Tokens such as ``pair of legs,'' ``tire,'' and ``hair'' denote accessory pieces commonly used in LEGO constructions, adding realism and complexity to assembly tasks. Recognizing and incorporating accessory pieces is essential for creating realistic and detailed models, and these tokens help the model understand the role of accessories in assembly.
    \item \textbf{Quality control and correctness.}
    Tokens like ``correct sticker'' and ``correct bag'' emphasize the importance of quality control and correctness in assembly tasks. Ensuring that the correct parts are used in the right context is essential for achieving accurate and high-quality assemblies, and these tokens highlight the need for attention to detail and accuracy in the assembly process.
    \item \textbf{Structural components.} 
    Tokens such as ``shaft,'' ``structure upside down,'' and ``roof tile'' suggest the inclusion of structural components and building techniques within the dataset. Understanding the role of structural components and mastering advanced building techniques is critical for creating stable and aesthetically pleasing assemblies, and these tokens provide guidance on constructing sturdy and well-balanced structures.
\end{itemize}
The theme entities included in our simulated dataset provide a realistic representation of the assembly tasks by encompassing part identification, spatial orientation, assembly techniques, component variations, accessory recognition, quality control, and structural components, the dataset contains the knowledge and skills necessary to effectively assemble LEGO structures in \ac{MR}.

\subsubsection{Conversation Tokens (Figure~\ref{fig:distribution_conversation})}

 We can infer several aspects that contribute to the usability and effectiveness of our simulated dataset for conversations during assembly training:
\begin{itemize}[nosep, leftmargin=*]
    \item \textbf{Role identification.}
    The presence of ``Trainer'' and ``Trainee'' tokens indicates a clear distinction between the roles of the instructor guiding the training session and the learner receiving instructions. This role identification fosters clarity and structure in the conversation, ensuring effective communication between trainer and trainee.
    \item \textbf{Instructional guidance.}
    Tokens such as ``step,'' ``plate,'' ``use,'' and ``find'' suggest the provision of instructional guidance within the conversation. The trainer entity likely provides step-by-step instructions and prompts to the trainee, guiding them through the assembly process and facilitating learning in a structured manner.
    \item \textbf{Interactive dialogue.}
    The conversation tokens include interactive dialogue cues such as ``Let's,'' ``Yes, let's,'' and ``Thank you!'' These cues foster engagement and collaboration between the trainer and trainee entities, creating a supportive and interactive learning atmosphere conducive to effective learning and skill development.
    \item \textbf{Feedback and encouragement.}
    Tokens like ``Great!'' and ``Alright'' suggest the inclusion of positive feedback and encouragement within the conversation. Positive reinforcement enhances motivation and engagement, encouraging active participation and fostering a positive learning experience for the trainee.
    \item \textbf{Error handling and assistance.}
    The presence of tokens like ``check,'' ``help,'' and ``completed'' indicates provisions for error handling and assistance within the conversation. The trainer entity likely offers guidance and support to the trainee in identifying and correcting errors, ensuring a constructive learning process and facilitating skill development.
    \item \textbf{Spatial orientation and task management.}
    Tokens such as ``right,'' ``left,'' ``back,'' and ``steps'' provide spatial orientation cues and references to assembly tasks. This spatial orientation facilitates effective communication of assembly instructions and task management between the trainer and trainee entities, ensuring accurate placement and alignment of LEGO components during assembly.
\end{itemize}
The conversation tokens provide instructional guidance, facilitate interactive dialogue, offer feedback and encouragement, handle errors, and provide spatial orientation cues for task management.

\subsubsection{Vision-language Tokens (Figure~\ref{fig:distribution_vlm})}
Analyzing the provided tokens from the vision language model, we can identify several factors contributing to its usability and effectiveness:
 
\begin{itemize}[nosep, leftmargin=*]
    \item \textbf{Object recognition.}
    Tokens such as ``plate,'' ``brick,'' ``tile,'' and ``knob'' represent common LEGO elements that users encounter during assembly tasks. By including these tokens, the dataset enables the vision language model to recognize and identify various LEGO components accurately, facilitating object recognition and understanding in \ac{XR} training environments.
    \item \textbf{Color detection.}
    Tokens like ``bright,'' ``grey,'' ``white,'' and ``blue'' provide color descriptors for different LEGO pieces. Incorporating color information allows the vision language model to detect and differentiate between LEGO components based on their color, enhancing the model's ability to interpret and analyze assembly scenes accurately.
    \item \textbf{Shape recognition.}
    Tokens such as ``round,'' ``flat,'' ``roof,'' and ``connector'' describe the shapes and configurations of LEGO elements. By including shape descriptors, the dataset enables the vision language model to recognize and classify different types of LEGO pieces based on their shapes, facilitating shape recognition and classification in \ac{XR} training environments.
    \item \textbf{Size specification.}
    Tokens like ``1x2,'' ``2x2,'' and ``1x1'' specify the sizes and dimensions of LEGO elements. Incorporating size information allows the vision language model to understand the scale and proportions of LEGO components within assembly scenes, aiding in size estimation and spatial reasoning during \ac{XR} training tasks.
    \item \textbf{Material and texture.}
    Tokens such as ``smooth,'' ``nougat,'' and ``transparent'' describe the materials and textures of LEGO elements. Including material and texture descriptors enables the vision language model to identify and distinguish between different surface finishes and textures, enhancing its ability to recognize and characterize LEGO components accurately.
    \item \textbf{Part relationships.}
    Tokens like ``side,'' ``top,'' and ``bottom'' provide spatial relationship cues between LEGO elements. By including part relationship descriptors, the dataset enables the vision language model to understand the spatial arrangement and orientation of LEGO components within assembly scenes, facilitating the interpretation of complex assembly structures and configurations.
    \item \textbf{Visual context understanding.}
    Tokens such as ``image'' and ``shows'' suggest the inclusion of visual context information within the dataset. Providing visual context cues enables the vision language model to interpret and analyze assembly scenes holistically, incorporating visual information to enhance its understanding of the surrounding environment and improve object recognition accuracy.
\end{itemize}
Our simulated dataset successfully provides object recognition, color detection, shape recognition, size specification, material and texture characterization, part relationships, and visual context understanding. 
Altogether,  these tokens contribute to the usability and effectiveness of the training environment by providing clear guidance, realistic representation of components and challenges, interactive dialogue, and enhanced vision understanding. These elements collectively enhance the learning experience and skill development in \ac{XR} assembly tasks. 

\begin{figure}[h]
\centering
\includegraphics[width=0.85\columnwidth, trim={0 0 0 0},clip]{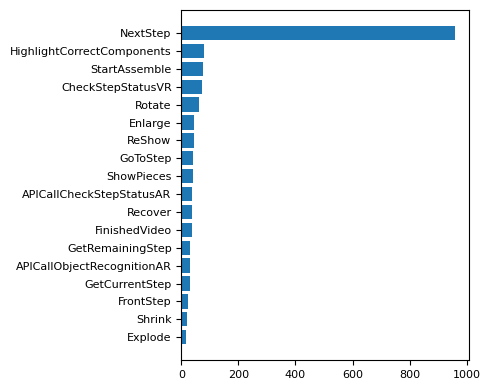}
\caption{Distribution of called tools in conversations.}
\label{fig:distribution_tool}
\end{figure}

\subsubsection{Called Tools (Figure~\ref{fig:distribution_tool})}

% [('NextStep', 959), ('HighlightCorrectComponents', 78), ('StartAssemble', 77), ('CheckStepStatusVR', 72), ('Rotate', 61), ('Enlarge', 46), ('ReShow', 45), ('GoToStep', 41), ('ShowPieces', 40), ('APICallCheckStepStatusAR', 39), ('Recover', 39), ('FinishedVideo', 37), ('GetRemainingStep', 32), ('APICallObjectRecognitionAR', 30), ('GetCurrentStep', 30), ('FrontStep', 22), ('Shrink', 19), ('Explode', 15)]
As shown in Figure~\ref{fig:distribution_tool}, we plot the distribution of the number of tools invoked in the generated conversations. 
The most frequently called and essential functional tools are those related to process control:
``NextStep'' (57.02\%), ``StartAssemble'' (4.58\%), ``CheckStepStatusVR'' (4.28\%), ``GoToStep'' (2.44\%), ``GetRemainingStep'' (1.90\%), ``GetCurrentStep'' (1.78\%), ``1.31\%''.
This indicates that users prioritize adherence to the assembly procedure during the fine-grained assembly task.
Functional tools related to user interactions are also significant, for example, ``HighlightCorrectComponents'' (4.64\%).

%%%%%%%%%%%%%%%%%%%%%%%%%%%%%%%%%%%%%%%%%%%%%%%%%%%%%%%%%%%%%%%%%%%%%%%% 3. Engineering details in the workflow
\subsection{Engineering Details in the Workflow}
\label{sec:appendix_lego}
\begin{figure}[h!]
\centering
\includegraphics[width=0.73\columnwidth]{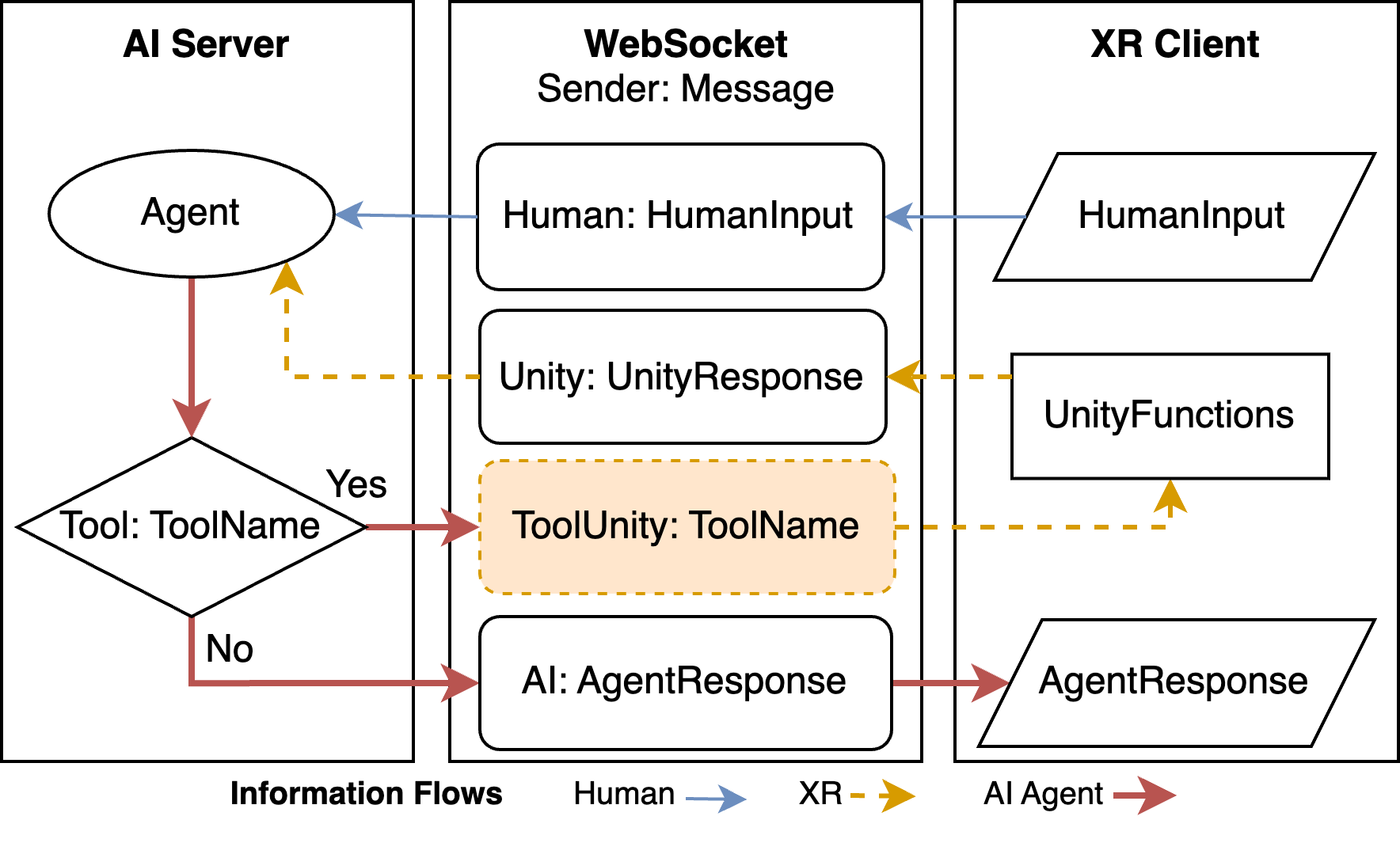}
\caption{Information flow.}\label{fig:information_flow}
\end{figure}

% \subsection{Information flow}
% %%%%%%%%%%%%%%%%%%%%%%%%%%%%%%%%%%%%
% \begin{figure*}[h]
% \centering
% \begin{minipage}[t]{0.48\textwidth}
% \centering
% \includegraphics[width=7.3cm]{figure/distribution_tool.png}
% \caption{Distribution of called tools in conversations.}
% \label{fig:distribution_tool}
% \end{minipage}
% \begin{minipage}[t]{0.48\textwidth}
% \centering
% \includegraphics[width=7.6cm]{figure/InformationFlows.png}
% \caption{Information flow.}\label{fig:information_flow}
% \end{minipage}
% \end{figure*}

%% file: table/prompt_template_and_task_instruction.tex
\begin{table*}[h!]
\resizebox{\linewidth}{!}{
\centering
\small
\begin{tabular}{@{}l@{}}
\toprule[1.5pt]
\textbf{(P1) Prompt template for user requirement generation} \\ \midrule
\begin{tabular}[c]{@{}l@{}}
{[}\textit{Task description}{]}\\ 
\begin{tabular}[c]{@{}l@{}}
You are an AI agent who acts as a Unity developer for AR applications.  
Your role is to analyze users' functional needs based on \\ the manuals and then develop the corresponding functions in an AR training system.
Note that is not for visually impaired users, \\ 
but for trainees who are visually healthy and able to wear HoloLen2 AR glasses.\end{tabular} \\ 
Here are samples of manuals:\\ 
{[}\textit{Manuals}{]}\end{tabular} \\ \midrule
\textbf{(P2) Prompt template for conversation generation}
\\ \midrule
\begin{tabular}[c]{@{}l@{}}
\underline{\textbf{1. System prompt~}} \\
{[}\textit{Task description}{]}\\
\textit{Brief version:}
The task is to generate multiple turns of conversations and called tools between the trainer (assistant) and  \\ trainee (user) grounded on the task-specific guidelines and tools in LEGO XR application. \\
\textit{Full version:}
The trainer aims to teach the trainee how to accomplish the 
assembly task based on the task-specific guidelines,\\ 
supported by an XR application. 
Specifically, the trainee is wearing AR glasses to see both VR environment and real world. \\
The trainee knows nothing about the guidelines before trainer's guidance.
For each step, the trainee must ask at least one deep-\\dive question, 
or request a troublesome issue if he or she cannot follow the guide,
or call tools from XR application and learn \\ how to use those tools;
the trainer must answer the question, assist the trainee,
show them the responses to the execution of the \\ tools.
At the end of a conversation,  first, the trainer must ask if the trainee has accomplished the task and the trainee must tell \\ if the trainee can accomplish the task; 
second, the trainer must ask how is user experiences, and the trainee provide feedback \\ on the user experience.
You must add a section title to separate which key point in the guideline 
in the generated conversation \\ and generate until the final step of the guidelines. \\
{[}\textit{Tool description as shown in Table~\ref{tab:tool_description}}{]}\\
\midrule
\underline{\textbf{2. Query prompt~}}\\
{[}\textit{Task description (Brief version)}{]} \\
{[}\textit{Summary and step instructions in a manual}{]}\\
Imagine some trainee's utterances have the intent of using the tools with the following responses: \\
{[}\textit{Tool responses}{]}\\
\end{tabular} 
\\ \midrule
\textbf{LEGO Assembly Assistant prompt (P3)} \\
\midrule
\begin{tabular}[c]{@{}l@{}}
You are a helpful AI assistant who aims to train the user how to assemble a LEGO car in XR immersive system.\\ Extended Reality (XR) directs to the assortment of Virtual Reality (VR), Augmented Reality (AR), and Mixed Reality (MR).\\ Please make sure you complete the objective above with the following rules:\\ (1) The user is a trainee who is wearing HoloLen 2 glasses and is able to see XR environments in real-time.\\ (2) You are able to call Unity functions in the LEGO AR application.\\ (3) You are able to obtain HoloLens 2 Sensor Streaming data.\\
(4) Alert if the user asks you something outside of the LEGO assembly task but do not give overconfident answers.\\ Your task is to answer the user's questions and assist the user in understanding how to complete the LEGO assembly task in XR.
\end{tabular}\\
\bottomrule[1.5pt]
\end{tabular}
}
\caption{Prompt templates used in this work.}
\label{tab:template_conversation_generation}
\end{table*}

%% file: table/user_requirements.tex
\begin{table*}[h!]
\centering
\renewcommand{\arraystretch}{1.15} % Default value: 1
\footnotesize
\begin{tabular}{lp{11.5cm}}
\toprule
User requirement & Description \\
\midrule
3D Model Interaction & Create 3D models of the LEGO pieces and the Monster Truck assembly. Trainees can interact with these 3D models using hand gestures and voice commands, making it easier to understand the assembly process. \\
Step-by-Step Guidance & Display step-by-step instructions directly in the trainees' field of view. This can include both visual instructions and written or spoken guidance. \\
Real-Time Feedback & Provide real-time feedback to trainees as they assemble the LEGO set. Use AR to highlight the correct attachment points and components, and indicate when they've completed a step correctly. \\
Object Recognition & Implement object recognition so that HoloLens 2 can identify LEGO pieces and highlight them when trainees look at them. This can help trainees quickly find the right pieces. \\
Progress Tracking & Keep track of trainees' progress and provide them with an overview of the steps they have completed and those remaining. This can help them stay organized and motivated. \\
Troubleshooting Assistance & Include a troubleshooting mode that guides trainees through common problems and solutions they might encounter during the assembly. \\
Data Logging & Collect data on trainees' performance and interaction with the AR training system to analyze their progress and make improvements to the training process. \\
\bottomrule
\end{tabular}
\caption{User requirements of the \ac{XR} training system.}
\label{tab:user_rerquirements}
\end{table*}